\documentclass[letterpaper, 10 pt, conference]{ieeeconf}  %

\IEEEoverridecommandlockouts                              %

\usepackage{hyperref}
\usepackage{graphics} %
\usepackage{epsfig} %
\usepackage{amsfonts} %
\usepackage{amsmath} %

\usepackage[T1]{fontenc}

\usepackage{xcolor}

\title{\LARGE \bf
Distributed Reinforcement Learning of Targeted Grasping with Active Vision for Mobile Manipulators
}

\author{
Yasuhiro Fujita,
Kota Uenishi,
Avinash Ummadisingu,
Prabhat Nagarajan,\\
Shimpei Masuda, and
Mario Ynocente Castro
\thanks{All authors are with Preferred Networks, Inc., Tokyo, Japan. {\tt\small [fujita, kota, ummavi, prabhat, masuda, marioyc]@preferred.jp}}%
}

\begin{document}

\bstctlcite{BSTcontrol}

\maketitle
\thispagestyle{empty}
\pagestyle{empty}

\begin{abstract}
Developing personal robots that can perform a diverse range of manipulation tasks in unstructured environments necessitates solving several challenges for robotic grasping systems.
We take a step towards this broader goal by presenting the first RL-based system, to our knowledge, for a mobile manipulator that can (a) achieve targeted grasping generalizing to unseen target objects, (b) learn complex grasping strategies for cluttered scenes with occluded objects, and (c) perform active vision through its movable wrist camera to better locate objects.
The system is informed of the desired target object in the form of a single, arbitrary-pose RGB image of that object, enabling the system to generalize to unseen objects without retraining.
To achieve such a system, we combine several advances in deep reinforcement learning and present a large-scale distributed training system using synchronous SGD that seamlessly scales to multi-node, multi-GPU infrastructure to make rapid prototyping easier.
We train and evaluate our system in a simulated environment, identify key components for improving performance, analyze its behaviors, and transfer to a real-world setup.

\end{abstract}

\section{Introduction}
\label{sec:intro}

In order to have robots bridge the gap from the laboratory to our homes, and to realize personal robots that can perform a diverse range of daily tasks and chores, such robots must be able to generalize across the vast diversity of unstructured domestic settings. In particular, such robots require the ability to perform \textit{targeted grasping} (also called instance grasping or target-oriented grasping), that is, to find and grasp target objects in cluttered scenes, across a diverse set of possibly unseen target objects.
Targeted grasping is in contrast to the \textit{untargeted grasping} (also called indiscriminate grasping) setting where robots are expected to grasp any visible objects from their workspace.
Cluttered scenes in the real world pose several challenges for such robots.
As the number of distractor objects increases, finding a target object from visual cues alone becomes more challenging.
When there are touching or occluding objects around a target object, a successful grasp requires the robot to execute complex strategies such as pushing, repositioning, and regrasping.
The partial observability induced by occlusion often necessitates a robot to actively move its vision sensor, known as \textit{active vision}.

Reinforcement Learning (RL) has been used for learning vision-based grasping~\cite{Boularias2015, vpg, kalashnikov2018qt, graspinginvisible}, as it enables self-supervised learning of complex grasping strategies including pregrasping behaviors.
In this paper, we take a step towards realizing the grasping capability required for personal robots by presenting the first RL-based system that can simultaneously (a) achieve targeted grasping generalizing to unseen target objects, (b) learn complex grasping strategies to address cluttered scenarios, and (c) moves its vision sensor to better locate target objects.
In contrast to prior work, we develop and evaluate our system on a mobile manipulator, as it can perform complex, non-localized tasks, which we expect for personal robots, and it permits active vision via moving its base and wrist camera.

\begin{figure}
\centering
  \includegraphics[width=1\linewidth]{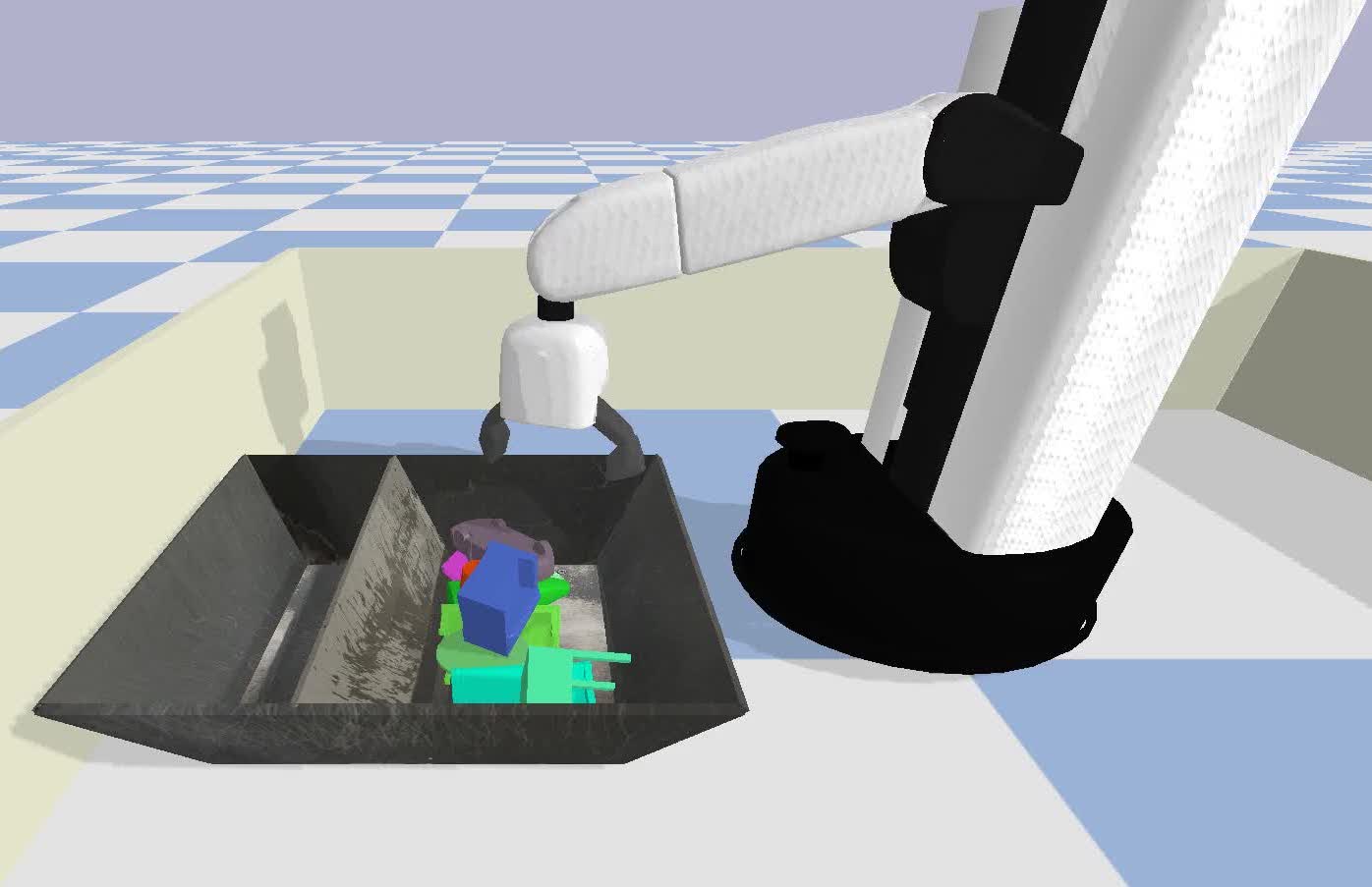}\\ \vspace{1ex}
  \includegraphics[width=.3\linewidth]{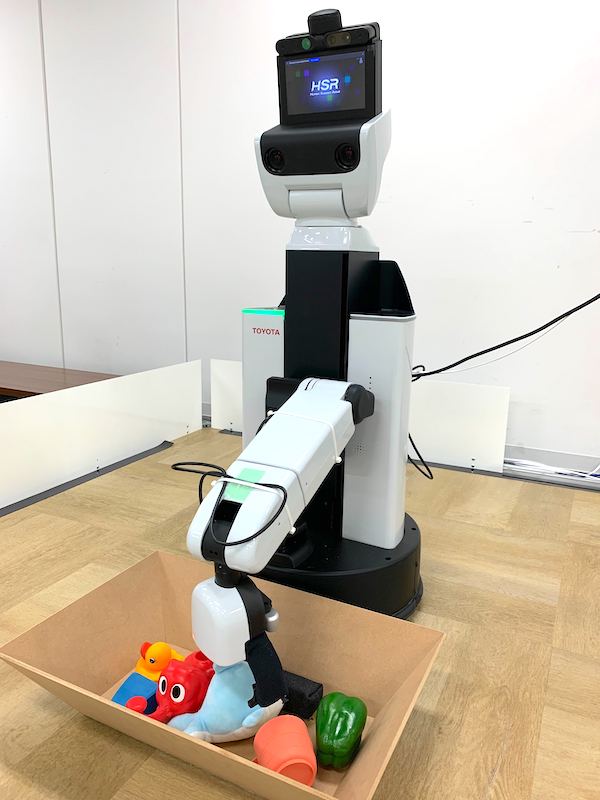}
  \includegraphics[width=.65\linewidth]{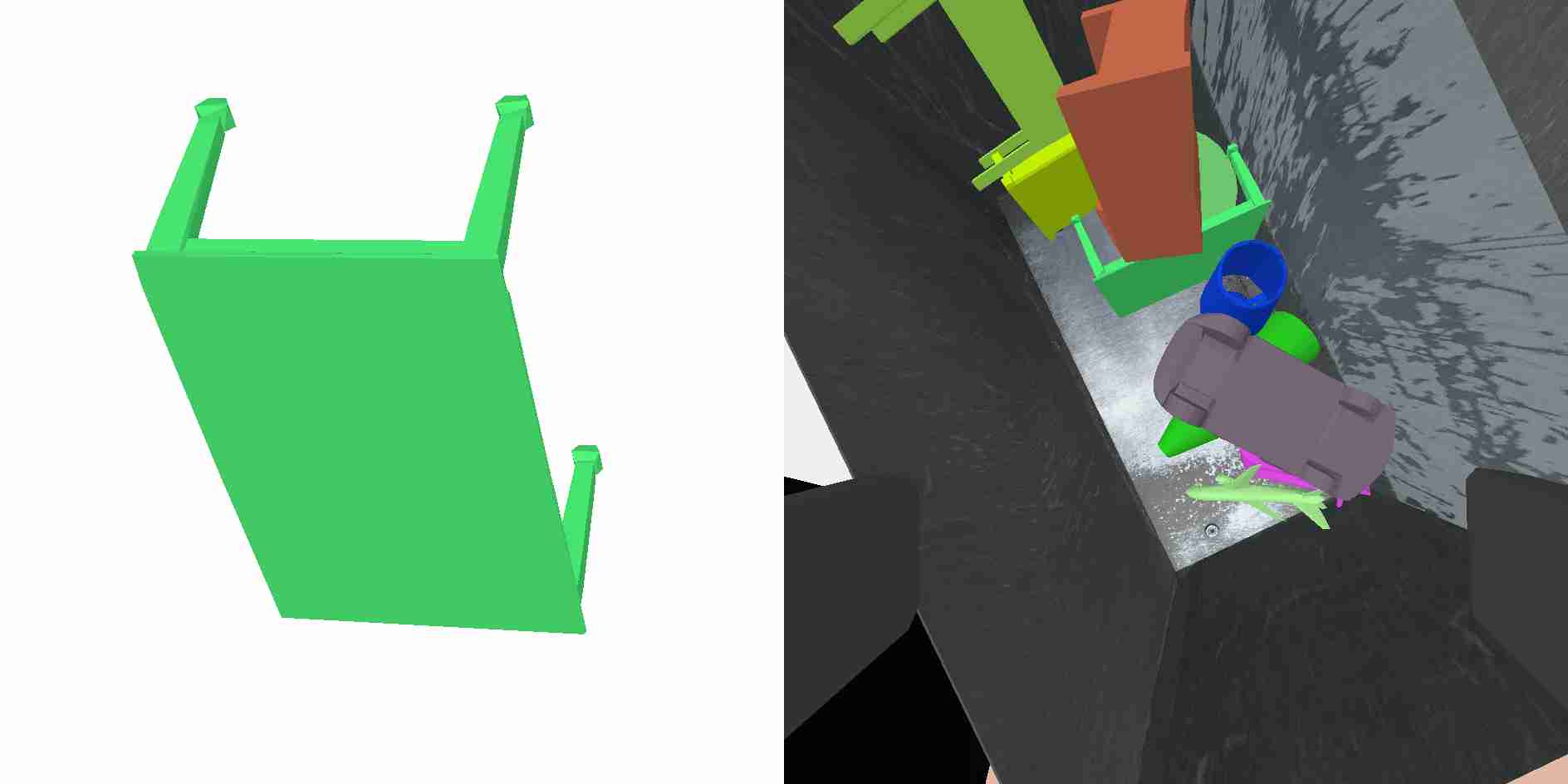}
  \caption{
\emph{top:} our simulated Human Support Robot (HSR) setup. The mobile manipulator must move its gripper and base to grasp a specific target object from a tray.
\emph{bottom-left:} our setup for real-world evaluation.
\emph{bottom-right:} example of the target image and an observation from the wrist camera given to the agent. The agent aims to produce a sequence of actions that will eventually grasp the given target object (the green table) from the tray. This may include pregrasping behaviors that position the object or the gripper to a more suitable orientation for a successful grasp.
}
  \label{fig:hsr}
\end{figure}

Accelerating computations through parallelism has been crucial for many recent successes of deep RL in challenging tasks~\cite{kalashnikov2018qt, alphastar, rubik, berner2019dota}.
Due to our task's large state and action spaces, sparse rewards, and partial observability, our system also requires an efficient distributed training architecture.
Thus, we develop such a system that significantly accelerates the training of our agent, which is characterized by large-batch synchronous stochastic gradient descent (SGD) with many asynchronous actors and distributed replay buffers.

The main contributions of our paper are as follows:
\begin{itemize}
    \item We present an RL-based system that enables a mobile manipulator to learn to successfully grasp unseen target objects in clutter through active vision via trial and error.
    Given an image of the target object, our system can learn to find and grasp the target object in a cluttered environment by moving the gripper and the camera jointly.
    We successfully transfer our learned policy to the real world and demonstrate targeted grasping on a physical mobile manipulator.
    \item We extend the QT-Opt algorithm \cite{kalashnikov2018qt} to the goal-conditioned setting, using hindsight experience replay (HER)~\cite{her}, frame stacking, data augmentation by mirroring, and a trick of lowerbounding target values to enable efficient learning.
    The effectiveness of each extension is measured in our ablation study.
    \item We present a distributed RL architecture that can scale up to 128 GPUs and 1024 CPU cores, with an effective batch size of 8192, accelerating the system to learn effective targeted grasping in 48 hours. 
    We also introduce a hyperparameter scaling trick for distributed adaptive gradient methods.
\end{itemize}

\section{Background}
\label{sec:background}

\subsection{Reinforcement Learning}

In \textit{reinforcement learning} (RL), a sequential decision-making task is formulated as a Markov decision process (MDP), which is a tuple $(\mathcal{S}, \mathcal{A}, T, \gamma, \mathcal{R})$, where $\mathcal{S}$ is the set of possible states, $\mathcal{A}$ is the set of possible actions, and $T$ is a transition function, which, given an action $a$, transitions an agent from a state $s$ to another, $s^{\prime}$.
At each state transition, the agent receives a reward $\mathcal{R}(s, a, s^{\prime})$.
The agent's objective is to learn a policy, a mapping $\mathcal{S}\rightarrow\mathcal{A}$ maximizing its expected return (or cumulative discounted reward) $\mathbb{E}[\sum_{t=1}^{T} \gamma^{t-1}\mathcal{R}(s_{t}, a, s_{t+1})]$, where $T$ is the length of the episode.
When the agent cannot observe the full information of the current state, the task is said to be \textit{partially observable}.

In \textit{goal-conditioned} RL, there is an additional goal-space $\mathcal{G}$, and the reward function is conditioned on a goal $g \in \mathcal{G}$, denoted $\mathcal{R}_{g}(s, a, s^{\prime})$.
The objective is to learn a goal-conditioned policy $\mathcal{S}\times\mathcal{G} \rightarrow \mathcal{A}$ maximizing $\mathbb{E}[\sum_{t=1}^{T} \gamma^{t-1} \mathcal{R}_{g}(s_{t}, a, s_{t+1})]$.

\subsection{QT-Opt}

Learning a policy that maximizes its expected cumulative discounted reward is often reduced to the problem of learning an action-value function $Q^{*}: \mathcal{S} \times \mathcal{A} \rightarrow \mathbb{R}$, where $Q^{*}(s,a)$ is the expected cumulative discounted reward the agent receives if it follows the optimal policy from a given state and action.

QT-Opt~\cite{kalashnikov2018qt} is an RL algorithm that learns $Q^{*}$ via function approximation.
QT-Opt stores experienced transitions of the form $(s, a, r, s^{\prime})$, that is, the agent's observed state $s$, the action $a$ taken in state $s$, the received reward $r$, as well as the state transition $s^{\prime}$, to a buffer called \textit{replay buffer}.
These transitions are uniformly sampled from the replay buffer to perform updates to approximate $Q^{*}$ by a neural network, following the update rule of deep Q-learning~\cite{mnih2015human}.
QT-Opt employs the cross-entropy method to maximize $Q^{*}(s,a)$ with respect to action $a$ to support a hybrid action space consisting of both discrete and continuous actions.
This enables a flexible action space supporting continuous pose changes and discrete gripper open/close movements, allowing pregrasping manipulation for robotic grasping.

\subsection{Distributed Training}

A distributed stochastic gradient descent (SGD) step consists of two inner steps:
(1) distributed workers calculate gradients from minibatches, and
(2) the model parameters are updated with those gradients.
These gradients can be aggregated either synchronously or asynchronously.
In \textit{synchronous SGD}, model parameters are updated synchronously with the average of the gradients of all workers.
On the other hand, in \textit{asynchronous SGD},
gradients from each worker are added to the model parameters on the
fly without synchronization.

Asynchronous SGD is fault-tolerant due to its nature of asynchrony where slow or failed workers can be ignored.
However, it also has the issue of stale gradients that introduces
nondeterminism to distributed training~\cite{atari21}. On the other hand, synchronous SGD
runs deterministically, making optimization more stable.
Although a naive implementation of synchronous calculation of the average of
gradients among workers can be slow due to tail
latency~\cite{tail40801, async45187}, it can be calculated very efficiently
by keeping the latency very low with HPC technology such as RDMA over InfiniBand.

For synchronous SGD, computing with $k$ processors (e.g. GPUs) in parallel makes the effective batch size $k$ times larger.
In large batch training, scaling the learning rate by $k$  
is known to work for momentum SGD~\cite{DBLP:journals/corr/GoyalDGNWKTJH17}.
However, in RL, adaptive learning rate methods such as Adam~\cite{kingma2014adam} have
been more widely used~\cite{atari21, AAAI1817184}, and the aforementioned scaling rule does not apply.
Lowering the hyperparameter $\epsilon$ by several magnitudes, e.g., from $10^{-3}$ to $10^{-8}$, is
known to work in large batch training~\cite{atari21}.

As the number of actors increases, the capacity of the replay buffer has to be increased as well to preserve the diversity of experiences. %
However, in general, scaling out a single data structure to multiple computers is itself a nontrivial task~\cite{dds2000}.  
The main difficulty stems from distributed consensus~\cite{flp10.1145/3149.214121} for consistency and tail
latency~\cite{tail40801} for performance.

\section{Problem Description and Formulation}
\label{sec:formulation}
\subsection{Problem Description} \label{sub:description}

In this paper, we address the problem of targeted grasping in clutter from RGB vision only, as depicted in Fig\@.~\ref{fig:hsr}.

We inform the robot of its target object via an independent RGB image of the target object, which we call a \textit{goal image}.
In contrast to approaches that rely on pretrained object detection or segmentation modules, our method does not need to predefine the set of possible target objects and is expected to generalize to unseen objects without retraining.
We do not assume the pose of the target object in the goal image to reduce the burden for users of the robot in preparing the goal image.

We consider a manipulator with a mobile base and a camera attached to its wrist to allow the robot to move the camera to better locate the target object.
The field of view of the wrist camera is limited (shown in the bottom-right image of Fig\@.~\ref{fig:hsr}), especially when the gripper is close to objects.
Moreover, dense clutter can cause significant occlusions, necessitating a grasping policy that leverages camera movement to overcome the issue of partial observability.

\subsection{Problem Formulation} \label{sub:formulation}

We formulate our problem as a goal-conditioned partially observable MDP as follows. 

\paragraph{Observation space}
At each timestep, the agent observes the RGB image from the wrist camera of the robot, a binary open/close state of the gripper, the $z$-position of the gripper, and the number of timesteps remaining in the episode (min-max normalized to the interval $[-1,1]$).
\paragraph{Goal space}
At the beginning of each episode, one of the objects is randomly selected as the target.
In addition to the agent's observations, the goal image is also fed to the agent as input in the form of an additional RGB image. 
\paragraph{Action space}
As in \cite{kalashnikov2018qt}, we allow two types of actions: a gripper pose change (3D position displacement and rotation along $z$-axis) and an open/close toggle command of the gripper. 
The gripper pose changes are done with respect to the camera's reference frame (at that timestep) as opposed to the global reference frame.
At each timestep, the agent must choose either type of action. 
While episodes terminate after a predefined maximum number of timesteps, there is an additional scripted termination condition~\cite{kalashnikov2018qt} to terminate an episode prematurely when the gripper is closed and lifted above a predefined height threshold. 
\paragraph{Reward function}
A reward of $1$ is given at the end of an episode if the robot successfully grasps the target object.
A small negative reward $-p$ is also given at each timestep to encourage faster task completion.
Formally, the goal-conditioned reward function is:
\begin{equation} \label{rewardeq}
    \mathcal{R}_{g}(s, a, s^{\prime}) = 
    \begin{cases} 
      1 - p & s^{\prime}=s_{term} \text{, } g \text{ grasped} \\
         -p & s^{\prime}=s_{term} \text{, } g \text{ not grasped} \\
         -p & s^{\prime} \neq s_{term} \\
  \end{cases}
\end{equation}
where $s_{term}$ denotes a terminal state, or the state at the end of an episode.

\section{System}
\label{sec:system}
\subsection{Learning Algorithm} \label{system_rl}

As our base learning algorithm, we selected QT-Opt~\cite{kalashnikov2018qt}, a variant of deep Q-learning~\cite{mnih2015human} that can learn in continuous action spaces and has demonstrated empirical successes in untargeted grasping.
As we demonstrate in Section~\ref{sec:experiments}, our targeted grasping task is sufficiently challenging that we must add multiple extensions to the algorithm in order to learn a successful policy.

\paragraph{Goal-conditioned QT-Opt}
To apply QT-Opt to the goal-conditioned setting, we augment the goal $g$ to the input, allowing the agent to learn $Q^{*}(s,a,g)$, the expected discounted reward the agent receives if it performs the optimal actions with respect to the desired goal. 
This follows the formulation of universal value function approximators (UVFAs)~\cite{uvfa}, which can learn to generalize to different goals and can be trained with Q-learning, as we do.

\paragraph{Hindsight experience replay (+H)}
While UVFAs in principle can learn values for different goals, 
UVFAs do nothing to remedy the sample complexity challenges posed by the sparse-reward nature of the targeted grasping task.
To alleviate the sample complexity issues posed by the targeted grasping problem, we replace QT-Opt's experience replay procedure with that of hindsight experience replay (HER)~\cite{her} to enable the agent to also learn from experiences where it successfully grasps an object other than the desired target object.
Suppose the agent's goal is to grasp the target object $g_{desired}$. 
However, instead, the agent grasps some other object $g_{achieved}$, where $g_{achieved} \neq g_{desired}$.
Traditionally, we would perform gradient updates to learn $Q(s, a, g_{desired})$.
However, with HER, we can replace the desired goal in hindsight, and reward the agent according to $\mathcal{R}_{g_{achieved}}(s, a, s^{\prime})$.
This allows us to update $Q(s, a, g_{achieved})$ according to reward $\mathcal{R}_{g_{achieved}}(s, a, s^{\prime})$, allowing the agent to receive a positive reward. 
This feedback \emph{rewards the agent for grasping a specific target object}, albeit not the one initially desired.
By doing so, we can de-sparsify the reward and allow the agent to receive positive rewards for successful grasps of any object, as is the case in the untargeted grasping setting.

\paragraph{Data augmentation via mirroring (+M)}
To further improve the utilization of the acquired experiences, we exploit the symmetry of the task for data augmentation~\cite{symmetry}.
This exploitation is enabled by the fact that the two-fingered gripper on the HSR is nearly symmetric, the fact that images from the wrist camera are also vertically symmetric (as seen in the bottom-right image of Fig\@.~\ref{fig:hsr}), and that the action space is aligned with the camera coordinates.
In particular, when a batch of transitions is replayed for gradient updates, for each transition, with probability 0.5, we reflect both the image observations and the actions.

\paragraph{Frame stacking (+F)}
To mitigate the partial observability of our task, we construct the input to the agent's neural network by stacking the last $N$ camera images from the wrist camera along the channel axis.
This technique, known as \emph{frame stacking}, is commonly used in RL for Atari 2600 games~\cite{mnih2015human}.

\paragraph{Lowerbounding target values (+L)}
Lastly, since every positive reward by a grasp success is delayed until the episode ends, assigning credit to actions from the reward is difficult, especially when learning values by bootstrapping the estimated value of the next state.
We introduce a simple trick, which is to lowerbound the target value of QT-Opt for a state by the discounted cumulative reward observed in the episode from that state\footnote{While this trick can lead to overestimation of Q-values under stochastic dynamics, we assume the dynamics of our task is sufficiently near-deterministic.
}.
When combined with HER, the discounted cumulative reward is replaced in hindsight as well.

\subsection{Distributed Training}

\begin{figure}[htbp]
  \centering
  \includegraphics[width=0.5\textwidth]{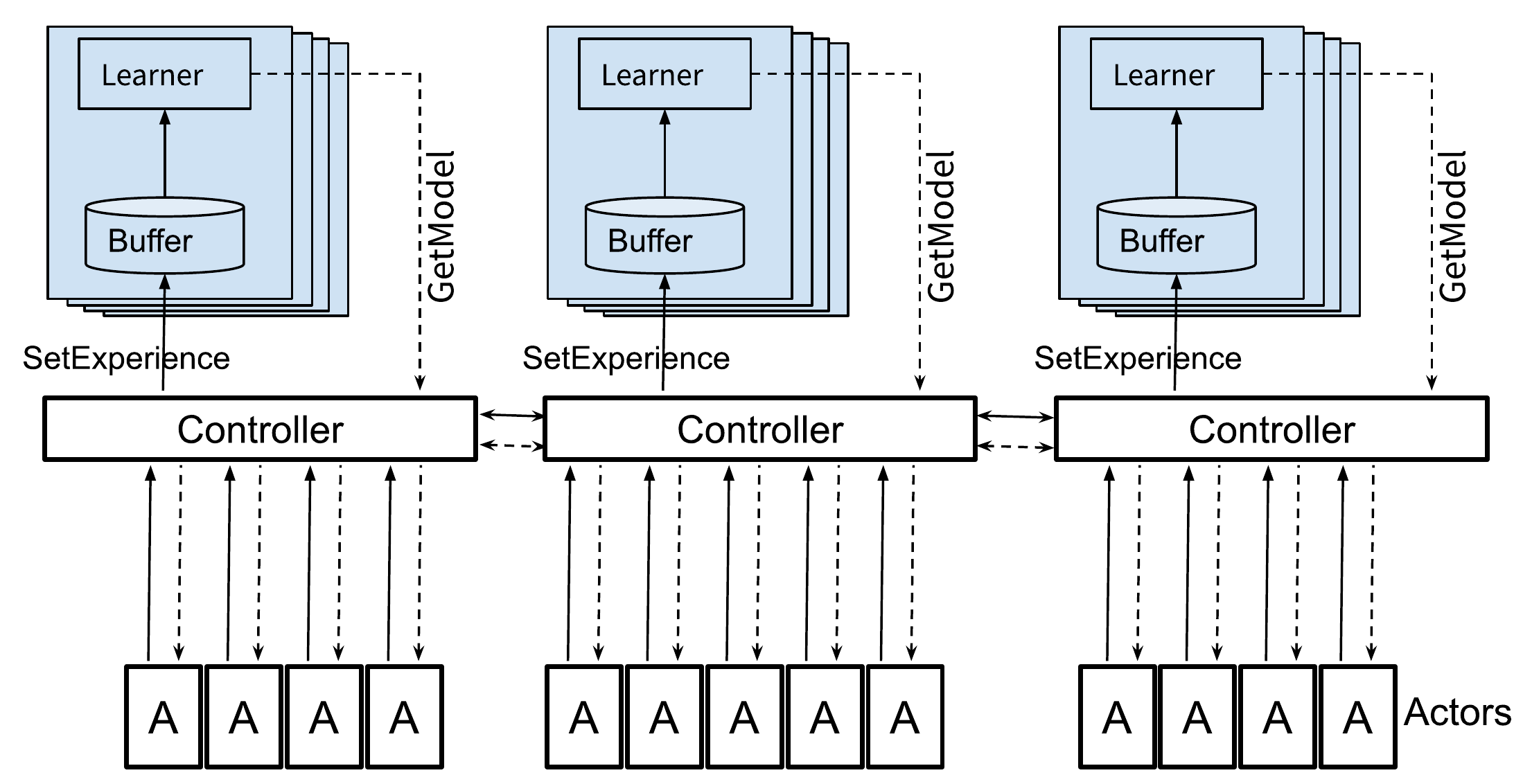}
  \caption{
An overview of our distributed training architecture with distributed replay buffers and remote actors.
A single node contains a single controller and multiple GPUs.
For each GPU in a node, there is a single learner with its own replay buffer. 
Replay buffers are in the process (address) space of learners to enable efficient sampling of minibatches.
Learners sample minibatches from their attached replay buffers, run distributed optimization on GPUs, and external remote actors fetch the model from learners through the controllers.
In practice, a node includes eight learners.
Actors append experiences to replay buffers via \texttt{SetExperience} RPC calls (solid-lined arrow) to learners.
Actors fetch model parameters from learners via \texttt{GetModel} RPC calls (dotted-lined arrow).
A controller is present in every node and balances RPC requests among learners.
}
  \label{fig:dist-replay-buf}
\end{figure}

Our distributed training architecture is depicted in Fig\@.~\ref{fig:dist-replay-buf}.
Following the nomenclature used by \cite{impala2018}, we refer to processes that run the optimization loop as \textit{learners}, and processes that collect experiences through environment interactions as \textit{actors}.
To further increase the utilization of experience collected, learners are distributed across multiple computer nodes, performing synchronous SGD by calculating the sum of gradients through a form of collective communication known as \textit{All-Reduce}.
Each learner has its own replay buffer to sample experiences from.
Each actor asynchronously collects and sends experiences into its assigned learner's replay buffer (\texttt{SetExperience}) and retrieves the latest model parameters from the learner (\texttt{GetModel}).
To cache the model parameters, we also implement processes named \textit{controllers} between actors and learners. They forward requests from actors to balance the load of learners as well as the number of experiences in replay buffers.

Providing each learner process with its own replay buffer is an essential design decision for our system.
Analogous to supervised learning, our method of distributing replay buffers across processes can be interpreted as local shuffling on a scattered training dataset, in contrast to global shuffling on a non-scattered training dataset.
Although global shuffling is known to lead to a solution slightly better than local shuffling in supervised learning~\cite{Meng2017ConvergenceAO},
we consider the effect small enough to trade-off for the performance and simplicity of our system by avoiding the difficulties of distributed systems.

\section{Experiments}
\label{sec:experiments}
\begin{figure}
\includegraphics[width=\linewidth]{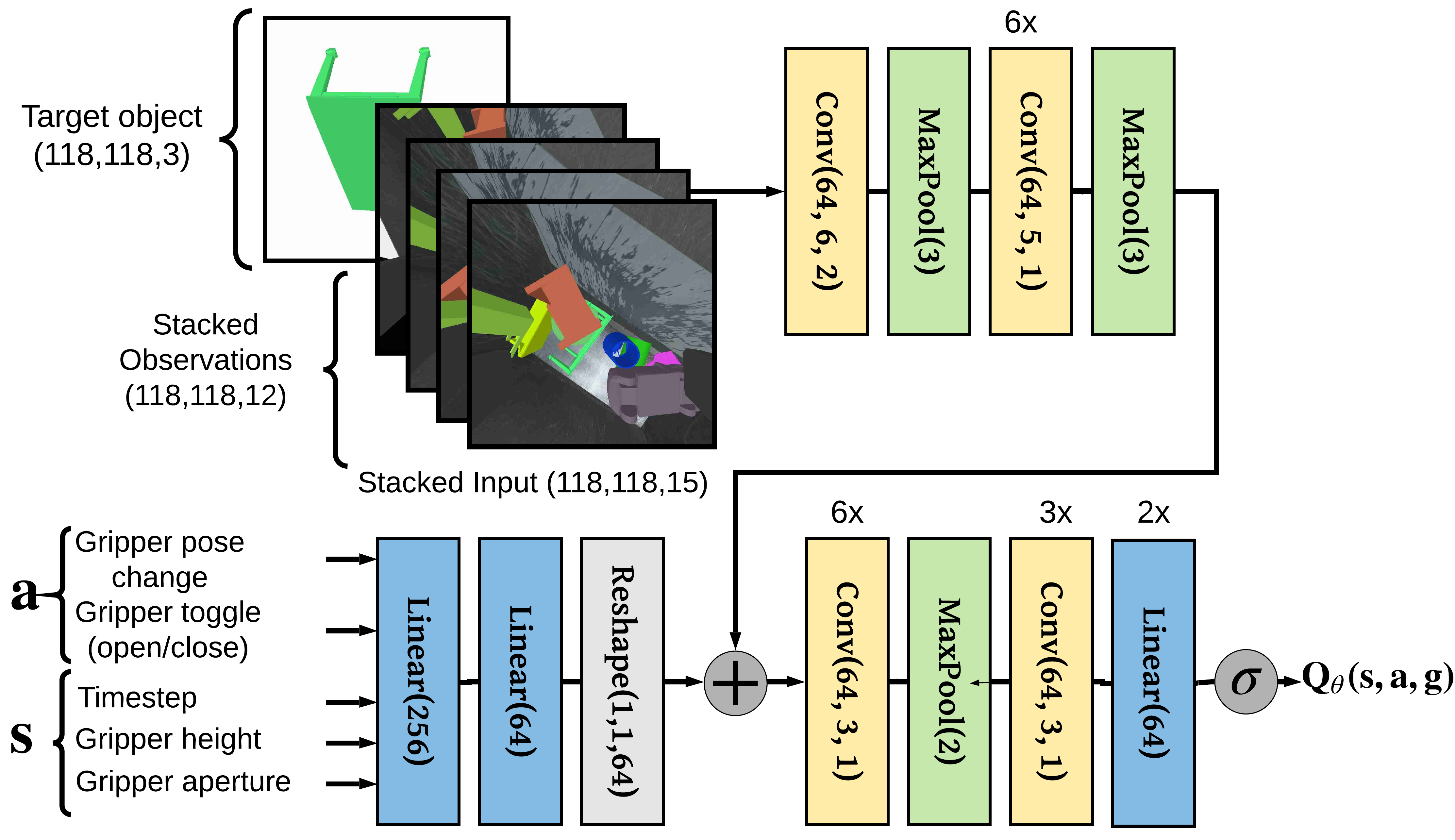}
\caption{
The neural network architecture we use for our experiments. 
Our network architecture is almost identical to the QT-Opt architecture~\cite{kalashnikov2018qt}, with the major differences being in the inputs. 
All convolution and linear layers are followed by batch normalization and a ReLU activation.
The input to the network is the previous four wrist camera observations and the goal image, all concatenated to form a 118x118x15 image input. In addition to the observation and goal, additional aspects of the state are fed to the network, including the number of timesteps (normalized to $[-1,1]$) remaining in the episode, the gripper height, and the gripper aperture. The action is also passed into the network, which includes the gripper pose change and the gripper toggle. The network outputs the predicted $Q$-value for the given observation, action, and goal (target object). In the untargeted grasping scenario, the network outputs $Q_{\theta}(s,a)$ and the goal image is not concatenated to the input.}
\label{fig:nn}
\end{figure}

\subsection{Simulated Environment}
\label{sub:simulator}
We developed a simulated environment for a Human Support Robot (HSR)~\cite{hsr}, a mobile manipulator equipped with a wrist camera.
We use the Bullet physics~\cite{coumans2019} engine to simulate the movement of an HSR robot and its interactions with objects. 
The HSR's movement along the $x$ and $y$ axes correspond to the movement of the HSR's base.

For candidate grasp objects, we use 3D object models from the ShapeNet dataset~\cite{shapenet2015}, a large-scale dataset of objects from a diverse set of semantic categories.
We also augment this data with a smaller set of randomly generated rigid objects~\cite{Quillen2018}.
For the ShapeNet objects, we follow the preprocessing steps laid out in \cite{graspgan}, of adjusting the objects' extents and masses, and randomly sample colors for the objects.
To avoid generating very thin objects that are nearly impossible for the HSR's gripper to grasp, we exclude 1051 objects (from 52467 ShapeNet objects) whose shortest axes are smaller than 5mm after adjusting their extents.
We split the objects into a training set (90\%) and a test set (10\%).
The test set is used exclusively to evaluate the agent.

At the beginning of an episode, we sample uniformly between 4 and 10 random objects and drop them in the tray to create a cluttered scene.
We randomly select one of these objects to be the agent's desired target object to grasp during the episode.

\subsection{Implementation Details}

We re-implemented the QT-Opt algorithm~\cite{kalashnikov2018qt} using ChainerRL~\cite{chainerrl}, and extended it to the goal-conditioned setting as described in section~\ref{system_rl}.
In HER, when replaying an experience, with 50\% probability, we use the original desired goal, unchanged.
The other 50\% of the time, we attempt to replace the desired goal with the grasped object.
If no object was grasped, the transition remains unchanged.

The agent uses $\epsilon$-greedy exploration starting from $\epsilon=1.0$, linearly decaying over 800K timesteps (cumulative timesteps across all learners) to an exploration of $\epsilon=0.2$, which remains constant for the remainder of the training. 

\begin{table}
    \caption{Hyperparameters values used for our experiments.}
    \label{hyperparams}
\centering
\begin{tabular}{c|c}
    \textbf{Hyperparameters} & \textbf{Value} \\ \hline
    \# stacked frames & 4 \\
    Adam alpha & 5e-5 \\
    Adam epsilon & 1e-2 / (effective batch size) \\
    \# evaluation episodes during training & 100 \\
    replay start size & 10,000 \\
    input image size & 118 x 118 \\
    replay buffer capacity & 1M \\
    effective batch size & 64 $\times$ \# learners \\
    min objects per episode & 4 \\
    max objects per episode & 10 \\
    discount factor & 0.9 \\
    maximum episode length & 20 \\
    per-timestep penalty $p$ & 0.05 \\
\end{tabular}
\end{table}

Our network architecture is shown in Fig.~\ref{fig:nn}. Table~\ref{hyperparams} summarizes the hyperparameter settings we use. 

\subsection{Distributed Training}

To assess the scalability of our training system, we trained models on the untargeted grasping task across several scales from 8 (1 node) to 128 GPU learners (16 nodes).
At all scales, 64 remote CPU actors are run remotely for each learner, as depicted in Fig.~\ref{fig:dist-replay-buf}.
The local minibatch size for each learner is 64, making the effective batch size $64k$, where $k$ denotes the total number of GPUs.
Additionally, the hyperparameter $\epsilon$ of Adam was scaled down to $\epsilon/64k$ by the effective batch size from $\epsilon=10^{-2}$.
For example, with the local batch size 64 per GPU and with 128 learners, the effective batch size is $8192$, and $\epsilon=10^{-2}/8192$.
One inspiration for our technique is the result by \cite{atari21}, that reducing $\epsilon$ by orders of magnitude can accelerate large minibatch training.

We ran distributed training in the untargeted grasping setting across different scales from 8 GPUs to 128 GPUs, as depicted in Fig.~\ref{fig:curves} (left).
The success rate of 128 GPU training with 1024 external CPU cores that run actors reaches 81\% in 3 hours 57 minutes and 96\% in 20 hours. On the other hand, at a smaller scale, it takes more than 23 hours for an 8 GPU job to reach 80\%, and 45 hours to reach 89\%.
Table~\ref{tbl:mn2} summarizes our cluster hardware specifications.

\begin{table}
    \caption{Hardware spec of a compute node we use.}
    \label{tbl:mn2}
    \centering
\begin{tabular}{c|c|c}
    CPU & Intel(R) Xeon(R) Gold 6254 (18C) & 2-way \\
    GPU & NVIDIA V100 (32GB) & x8 \\
    Memory & 384GB DDR4-2933 & \\
    Network & Mellanox CX4 100Gbps Ethernet & x4
\end{tabular}
\end{table}

\subsection{Performance Evaluation}

\begin{figure}
\centering
\includegraphics[width=\hsize]{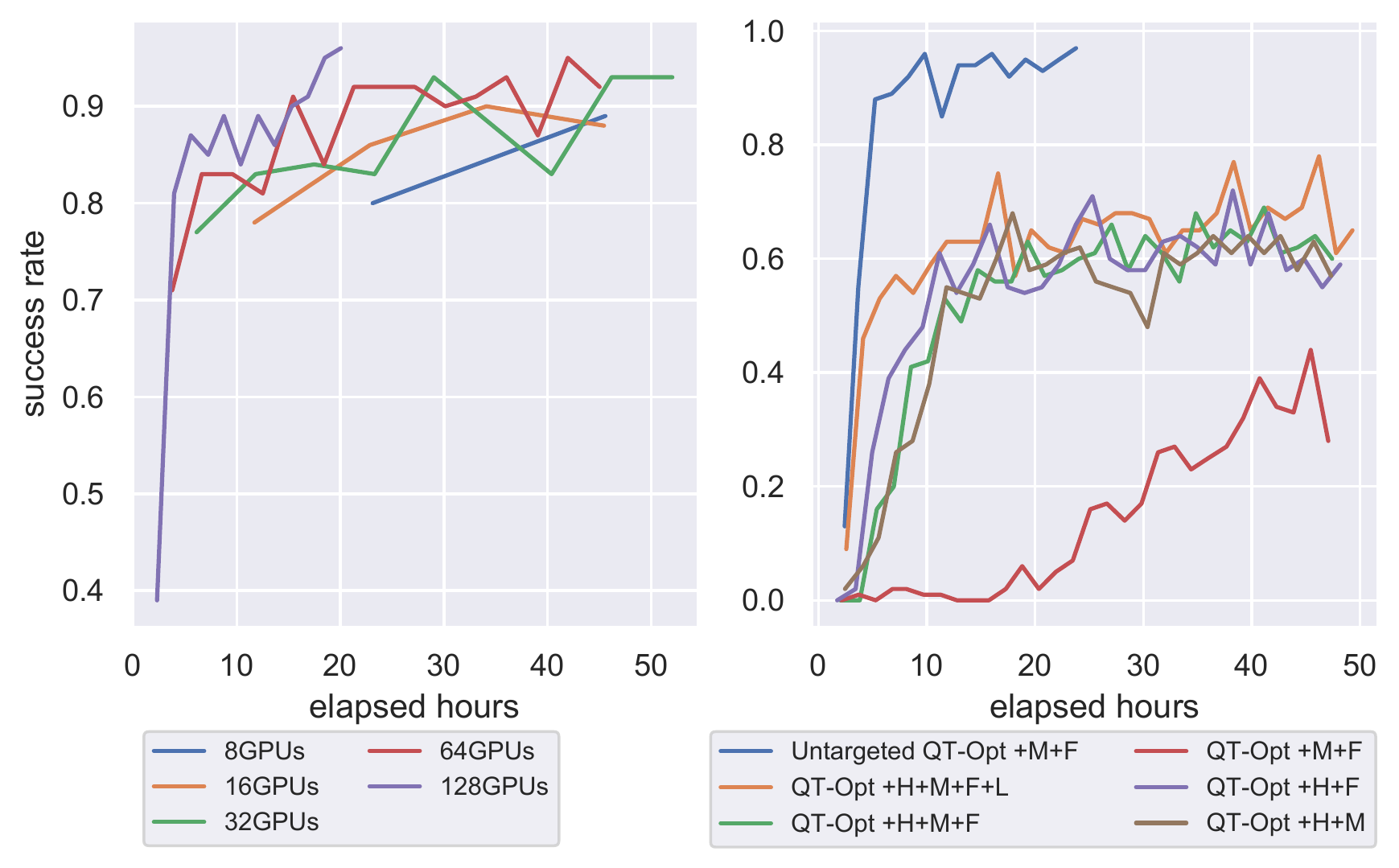}
\caption{
Grasping success rates during training for different configurations.
\emph{left:}
Untargeted grasping with a varying number of GPUs.
As the number of GPUs increase, the success rate of evaluation episodes grows faster.
\emph{right:}
Four extensions to QT-Opt are evaluated with 64 GPUs: HER (H), mirroring (M), frame stacking (F), and target lowerbounding (L).
The targeted grasping setting is used unless specified otherwise.
}
\label{fig:curves}
\end{figure}

We evaluate our system on both the untargeted and targeted settings.
In the targeted setting, we evaluate the proposed extensions to QT-Opt: HER, mirroring, frame stacking, and value lowerbounding.
Fig.~\ref{fig:curves} (right) shows the training curves, where each data point corresponds to the evaluation of a model for 100 test episodes.
Among the proposed extensions to QT-Opt, HER and value lowerbounding are the key extensions that sufficiently accelerate training to make training within two days possible.
Data augmentation with mirroring and frame stacking do not make a noticeable difference in success rates in this setting.
We also see that the targeted setting is significantly more challenging than the untargeted setting, not only in terms of the training speed but also in terms of asymptotic grasping success rate.

\begin{table}
    \caption{
Grasping success rates with 1000 test episodes.
}
    \label{tbl:final}
    \centering
\begin{tabular}{c|c|c}
 & ShapeNet and random & random only \\ \hline
    QT-Opt +M+F Untargeted & 94.4\% & 93.9\% \\
    QT-Opt +H+M+F+L Targeted & 66.3\% & 79.0\% \\
\end{tabular}
\end{table}
Table~\ref{tbl:final} shows the grasping success rates of the best model found during training, re-evaluated for 1000 test episodes.
The ``ShapeNet and random'' column shows the success rates with both the  ShapeNet and randomly generated objects, while the ``random only'' column shows results when the test objects are only randomly generated objects.
Our system can learn to grasp a target object successfully in 66.3\% of the test episodes in two days of training.
In the untargeted setting, our system can achieve a success rate of 94.4\%, which is consistent with the reported results in similar untargeted settings~\cite{kalashnikov2018qt}.
If we restrict the test objects to the randomly generated objects, the success rate of targeted grasping significantly increases, suggesting that the ShapeNet objects are significantly more difficult for the system.

\subsection{Qualitative Analysis}

\begin{figure*}
\centering
\includegraphics[width=\hsize]{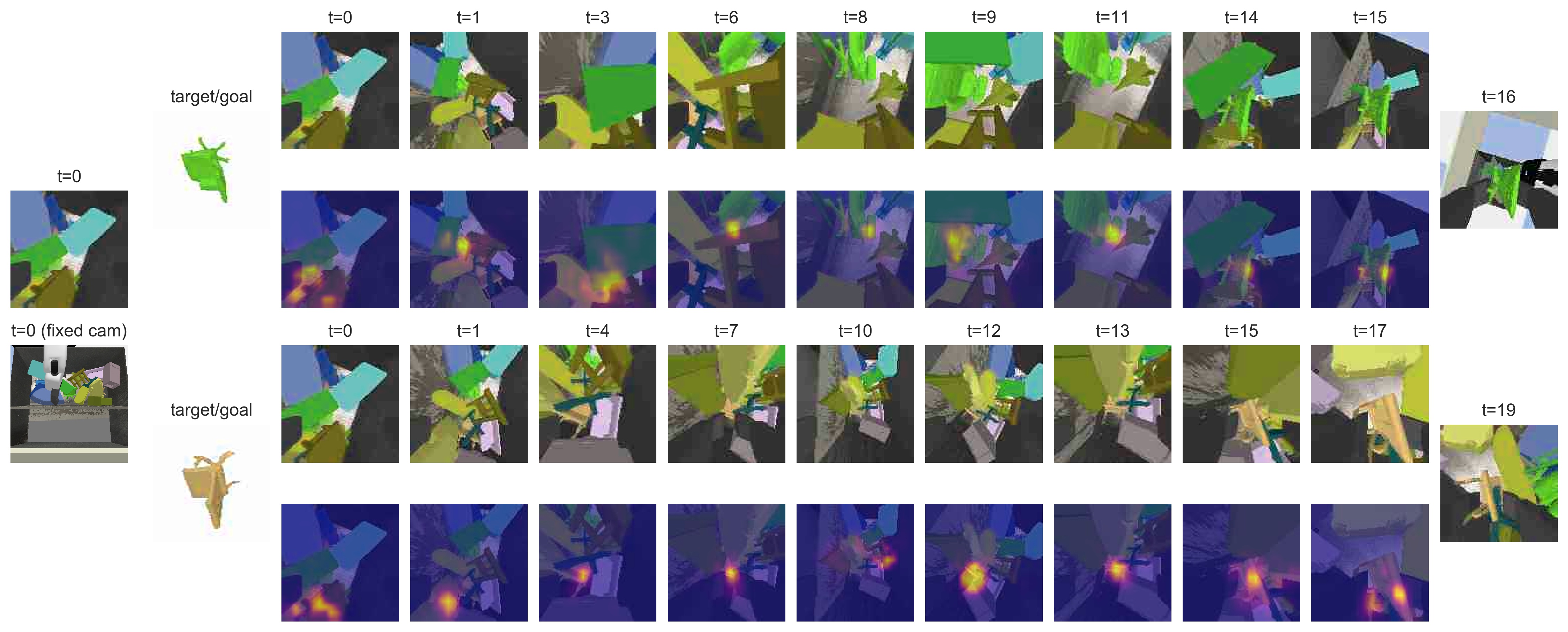}
\caption{
Sensitivity analysis of two challenging grasp attempts using a trained QT-Opt +H+M+F+L in clutter (15 objects). The leftmost frame depicts the first observation from which both trajectories begin, as well as a corresponding view from an external camera (visualization only). The first row shows the progression of the grasp attempt for the green chair. From the top-view, it is clear that the target is obscured by a green table, requiring the agent to perform a series of emergent pregrasp manipulation behaviors including moving away for better visibility ($t=1$), moving objects around the target's neighborhood away ($t=2$ to $t=8$), nudging the obstructing table aside ($t=9$), and finally grasping the object. The second row shows, overlaid on the observation, its sensitivity map. This highlights regions of the observation that, when perturbed, affect the predicted Q-values. This allows for more insight into these emergent behaviors, such as target localization ($t=1$ onwards), grasp region prediction ($t=9$), grasp success ($t=15$), etc.
The third row shows the grasp attempt for the beige-colored chair with the fourth showing its corresponding sensitivity map. As seen in the initial observation ($t=0$), the chair is not present in the agent's limited view and must first be found ($t=1$), despite potentially being fully occluded in cases of dense clutter or  large objects. Once located, the agent attempts to grasp it and fails ($t=7$). It then attempts re-grasping from a different angle and eventually succeeds.  
}
\label{fig:sensitivity}
\end{figure*}

\begin{figure}
\centering
\includegraphics[width=\hsize]{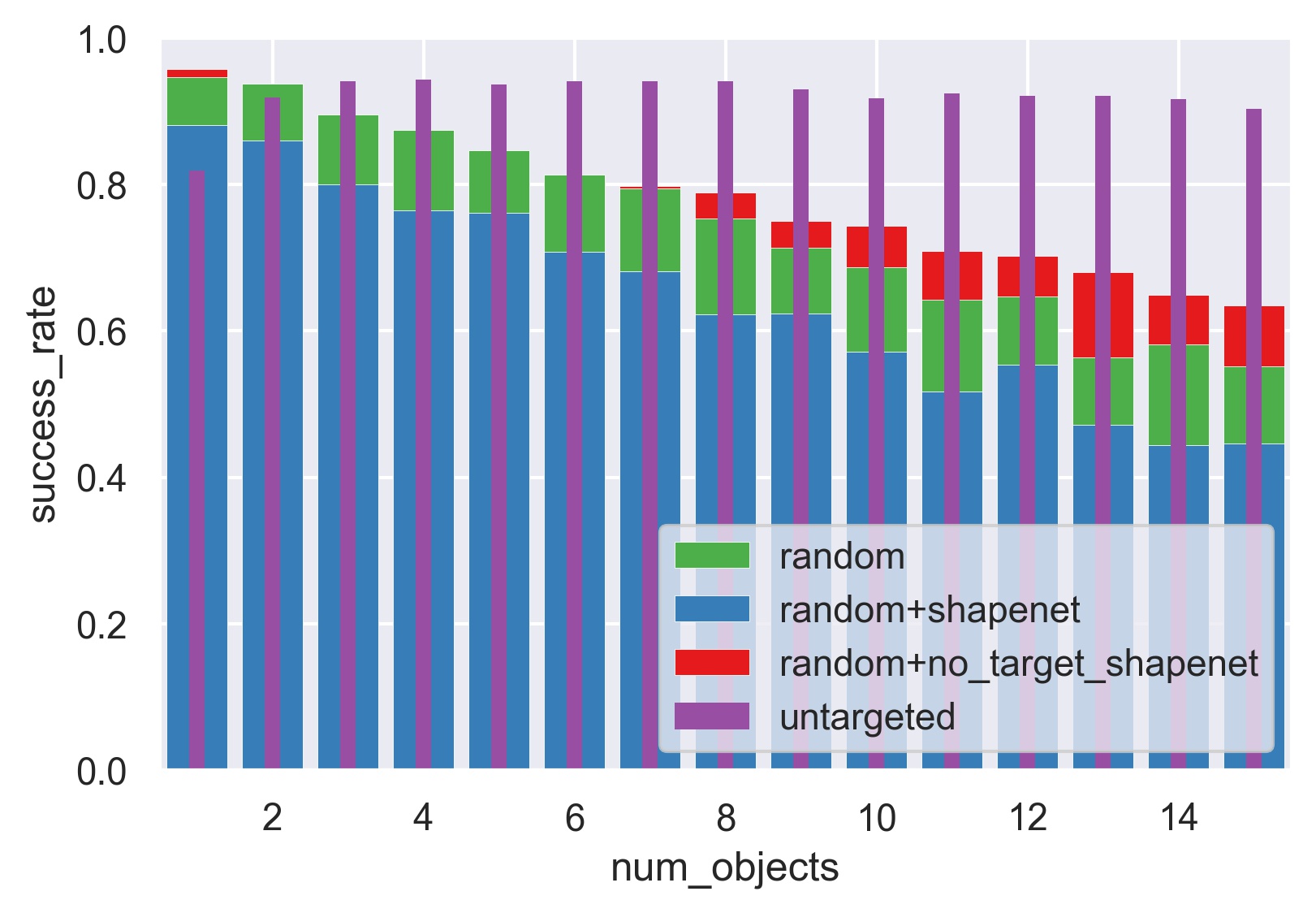}
\caption{
Grasp success rate of QT-Opt +H+M+F+L evaluated with increasing numbers of objects at evaluation time. \textit{random} was evaluated on the randomly generated rigid objects~\cite{Quillen2018}. \textit{random+shapenet} includes objects from the ShapeNet dataset~\cite{shapenet2015} and is the combination of datasets trained on. \textit{random+no\_target\_shapenet} is similar to \textit{random+shapenet}, but does not use ShapeNet objects as a grasp target.
}
\label{fig:num_objects}
\end{figure}
In order to better understand the learned grasping behaviors, we perform additional investigations using the trained QT-Opt +H+M+F+L model described in the previous sections.

\emph{Sensitivity Analysis:} Fig\@.~\ref{fig:sensitivity} depicts two grasp attempts from the same initial configuration (shown left-most), under two different desired target objects to grasp. The first row that follows depicts a sequence of observations as the agent attempts to grasp the green chair. Each timestep has a corresponding sensitivity map below it that measures areas of the observation, which, when perturbed, affects the predicted Q-values~\cite{greydanus2018visualizing}. This analysis enables the visualization of regions of the observation deemed as most important for the success/failure of the grasp by the agent's neural network. In the sensitivity maps early in the episode (around $t=1$), the regions with the highest sensitivity (orange) appear to correspond to the location of the target object in the scene, if present in the scene, despite being heavily occluded. This suggests that the agent has implicitly learned \emph{multi-scale localization} of the target object in the scene without any explicit formulation or use of a separate module for semantic segmentation or localization. As the agent's gripper moves closer to the object and begins a grasp attempt, the maps tend to show two or more regions of activity around the perimeter of the target object (around $t=15$). These regions typically correspond to areas where the agent's gripper will proceed to make contact with the object, thereby suggesting an implicit form of \emph{grasp prediction}. This analysis also lends insight into more emergent pregrasping behaviors, such as seen at $t=3$ for grasping the green chair. The target object is completely occluded in the observation as the agent moves away nearby objects to make space for the grasp. The sensitivity map suggests that the agent keeps track of the target's expected position despite being completely occluded, thereby directing its pregrasping behavior. Further evidence of this is seen at $t=6$ where the region around the brown table's leg is highlighted before the agent proceeds to push it out of the way to clear a path to the object. 
These goal-oriented pregrasp behaviors displayed by the agent are not explicitly encouraged in training and emerge solely as a consequence of trial-and-error using the success/failure reward signal.

\emph{Grasp target analysis:} Fig\@.~\ref{fig:num_objects} shows how our agent performs under varying degrees of clutter and with different types of objects. Despite being trained with between 4 and 10 objects sampled from the \emph{random} and \emph{ShapeNet} datasets, we see that the agent (QT-Opt +H+M+F+L) is able to achieve a grasp success rate of 44.6\% in the highly challenging setting of 15 objects under significant clutter and occlusion with \emph{random+shapenet} objects. Investigation revealed that grasping of some objects in the highly diverse ShapeNet dataset is infeasible due to physical limitations of the gripper, despite steps taken to eliminate objects that are nearly impossible to grasp. In order to discount its effect, we evaluate the agent with objects from only \emph{random}, yielding an impressive 15-object success rate of 55.1\%. Due to the visual similarity of models in the \emph{random} dataset, especially in the cluttered setting, we also evaluate the agents by re-introducing ShapeNet objects, while disallowing them from being the grasp targets. The QT-Opt +H+M+F+L evaluated using \emph{random+no\_target\_shapenet} achieves a remarkable success rate of 63.5\% in the difficult cluttered setting of 15 objects, thereby substantiating the efficacy of our method.

\subsection{Real-World Evaluation}

\begin{figure}
\centering
\includegraphics[width=0.5\hsize]{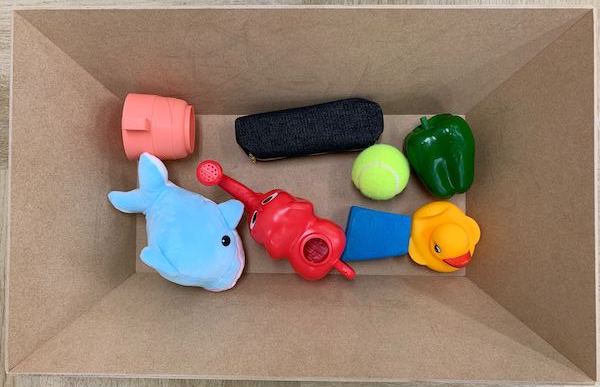}
\includegraphics[width=0.32\hsize]{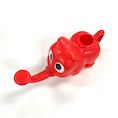}
\includegraphics[width=0.3\hsize]{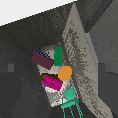}
\includegraphics[width=0.3\hsize]{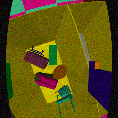}
\includegraphics[width=0.3\hsize]{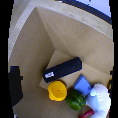}
\caption{
\emph{top left:} The tray and the eight objects used for the real-world experiment.
\emph{top right:} A goal image (photographed on a white desk) provided to the real HSR.
\emph{bottom-left:} Simulator images with and without domain randomization.
\emph{bottom-right:} A real HSR's wrist camera observation.
}
\label{fig:realexperiment}
\end{figure}

\begin{figure}
\centering
\includegraphics[width=\hsize]{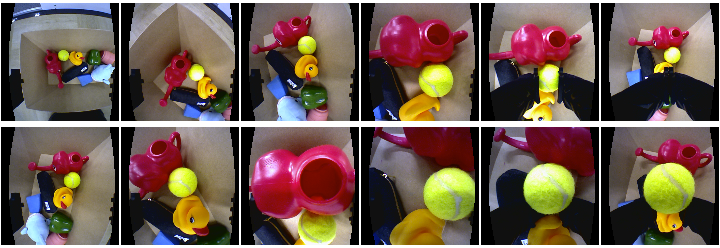}
\caption{
A sequence of wrist camera observations from a successful attempt to grasp the tennis ball.
}
\label{fig:real_traj}
\end{figure}

Lastly, we conduct an experiment to assess the real-world transferability of a model trained using our system to the setup depicted in the bottom-left of Fig.~\ref{fig:hsr}.
We finetune the image processing component of the model (corresponding to the top portion of Fig.~\ref{fig:nn}) using a dataset of simulated scenes with and without randomization of texture, color, lighting, and camera parameters~\cite{domainrandomization}, as shown in Fig.~\ref{fig:realexperiment}. We do this by matching the activations of the component's final layer for non-domain-randomized inputs to those of domain-randomized inputs by minimizing their mean squared errors.
Using a tray and eight different objects depicted in Fig.~\ref{fig:realexperiment}, we evaluate five targeted grasp attempts for each object.
Out of the 40 trials, our system succeeds in targeted grasping 25 times.
Fig.~\ref{fig:real_traj}, depicting one of the trials, shows a successful grasp of a tennis ball.
The most common failure scenario is a grasp of a distractor object instead of the desired target object, which constitutes 11 of the 15 failures. 
A video of the experiment is available at \url{https://www.youtube.com/watch?v=7RdLxAgwOAY}.

\section{Related Work}
\label{sec:relatedwork}

The task of targeted grasping in clutter has been the subject of investigation of multiple works~\cite{danielczuk2019mechanical,zeng2018robotic,Quillen2018,graspinginvisible,grasp2vec}.
Danielczuk et al.~\cite{danielczuk2019mechanical} address a similar problem to ours, but they rely on a separate perception module and a heuristic policy for pushing.
Zeng et al.~\cite{zeng2018robotic} develop a pick-and-place system that can grasp objects in clutter, but without pregrasping behaviors.
Both works are similar to ours in that they represent goals through images of the target object.
Quillen et al.~\cite{Quillen2018} benchmark several vision-based RL algorithms for targeted grasping, where the goal is to grasp objects of a specific shape as opposed to arbitrary unknown objects.
Yang et al.~\cite{graspinginvisible} also employ RL to tackle a related problem of ``grasping the invisible'', where the target object is completely occluded.
Unlike our system, which takes in the target object as an image, they rely on a segmentation module that must be able to detect the target object, restricting their system to objects detectable by the module.
Grasp2Vec~\cite{grasp2vec} uses object grasps to learn representations for objects, to provide self-supervision for goal-conditional grasping. 
However, one drawback of their method is its dependence upon an over-the-shoulder camera to learn object-centric representations.
Ebert et al.~\cite{ebert2018robustness} develop a system that can learn complex manipulation behaviors but do not study the problem of targeted grasping, but rather targeted configuration, with goals being target configurations instead of target objects.
None of the aforementioned RL-based systems perform targeted grasping in the active vision context, as they use either overhead, over-the-shoulder, or fixed mounted cameras and cannot alter the camera position.

Active vision systems have been developed for target or goal-driven tasks, especially in navigation, where the agent must reach a target~\cite{mousavian2019visual,zhu2017target}. 
Active vision has also been used extensively to address the problem of object perception~\cite{wu2015active,novkovic2019object,jiang2016novel}, a key subtask of targeted grasping, but it is less commonly applied to grasping and manipulation problems in clutter~\cite{pmlr-v87-cheng18a, zeng2018robotic}.
Perhaps the most closely related work to ours is Cheng et al.'s work ~\cite{pmlr-v87-cheng18a}, which uses RL for active vision on a goal-directed pushing task, where an object must be pushed to a target location in the presence of distractor objects.

Several frameworks have been proposed to leverage distributed computing in RL.
Gorila~\cite{nair2015massively} is a distributed training architecture for DQN~\cite{mnih2015human} that updates parameters asynchronously, while IMPALA~\cite{impala2018} relies on synchronous updates.
Both consider the case where there is a single, global replay buffer, and actors are run remotely.
Stooke et al.~\cite{stooke2019rlpyt} have developed a codebase of a wide range of RL algorithms with distributed training support with synchronous SGD and asynchronous (but local) actors.
Adamski et al.~\cite{atari21} have investigated the use of large minibatches in deep RL with synchronous SGD, but their work is limited to on-policy methods without experience replay.

\section{Conclusion}
\label{sec:conclusion}
We have presented a distributed deep reinforcement learning system for targeted grasping with active vision that can grasp unseen objects in dense clutter.
We have shown that some of our proposed extensions and the large-scale distributed training are key to learning efficiently on this challenging task.

While we also have shown a proof-of-concept demonstration of real-world transfer, simulation-to-real (sim2real) transfer of a vision-based manipulation system is a difficult challenge in itself.
We expect our system can benefit from recent advances in sim2real transfer to improve its performance in the real-world setup.

\addtolength{\textheight}{-5cm}   %

\section*{Acknowledgment}
We thank Koichi Ikeda, Kunihiro Iwamoto, and Takashi Yamamoto from Toyota Motor Corporation for their support of the HSR robots, and Kentaro Imajo for his help in designing and implementing the distributed training software.

\bibliographystyle{IEEEtran}
\bibliography{IEEEabrv,citations}

\end{document}